\documentclass{article}

\usepackage[preprint]{neurips_2022}

\usepackage[utf8]{inputenc} 
\usepackage[T1]{fontenc}    
\usepackage{hyperref}       
\usepackage{url}            
\usepackage{booktabs}       
\usepackage{amsfonts}       
\usepackage{nicefrac}       
\usepackage{microtype}      
\usepackage{xcolor}         
\usepackage[pdftex]{graphicx}

\title{Tag Embedding and Well-defined Intermediate Representation improve Auto-Formulation of Problem Description}

\author{%
  Sanghwan Jang \\
  Department of Computer Science and Engineering \\
  POSTECH \\
  Pohang, Korea \\
  \texttt{s.jang@postech.ac.kr} \\
}

\begin{document}

\maketitle

\begin{abstract}
  In this report, I address auto-formulation of problem description, the task of converting an optimization problem into a canonical representation. I first simplify the auto-formulation task by defining an intermediate representation, then introduce entity tag embedding to utilize a given entity tag information. The ablation study demonstrate the effectiveness of the proposed method, which finally took second place in NeurIPS 2022 NL4Opt competition subtask 2.
\end{abstract}

\section{Auto-Formulation of Problem Description}

\begin{figure}[ht!]
  \centering
  \includegraphics[width=1.0\linewidth]{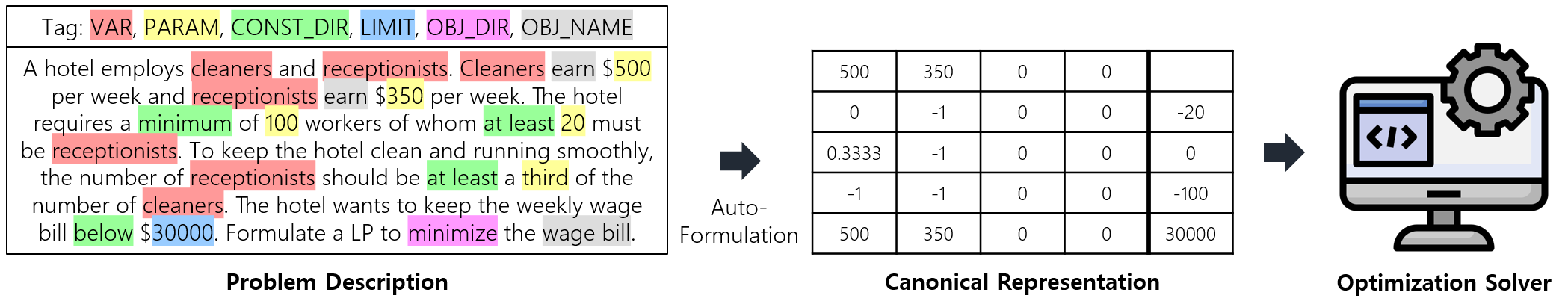}
  \caption{Auto-formulation of problem description.}
\end{figure}

Auto-formulation is the task of converting an optimization problem described in natural language into a canonical representation that an optimization solver can process. In this study, a problem description and tagged entities such as variables and constraint directions are given, and the proposed method should extract the coefficients and constants of the objective and constraints for the given linear programming problem.

\section{Two-Stage Auto-Formulation}

\begin{figure}[ht!]
  \centering
  \includegraphics[width=1.0\linewidth]{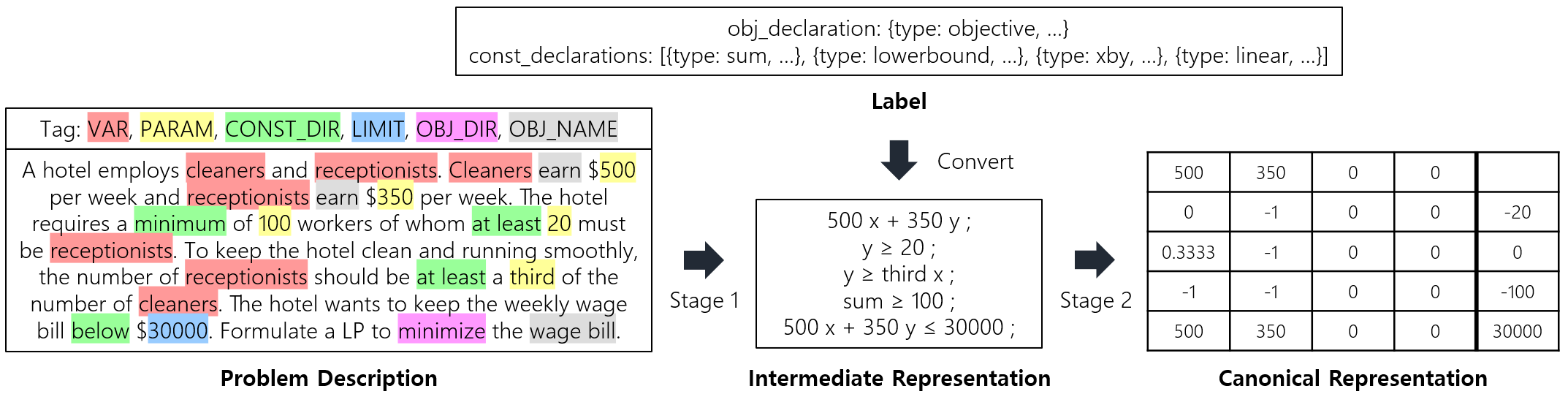}
  \caption{Two-stage auto-formulation.}
\end{figure}

I decompose auto-formulation task into two stages: 1) translating an optimization problem into an intermediate representation and 2) converting the intermediate representation into a canonical representation. This allows the model to address only the first stage, which eases the training of the model.

\subsection{Intermediate Representation}

I define intermediate representation to be in the form of mathematical expression (e.g., $3x + 4y \leq 50$), which pretrained BART already knows. The objective and constraints are generated at once so that the model can auto-formulate an optimization problem considering the relationship between declarations. To avoid inconsistency in model training, I define the generation order of declarations and convert labels to intermediate representations accordingly. The rules for the generation order are:

\begin{enumerate}
  \item For declarations: objective $\rightarrow$ constraints
  \item For constraints: lowerbound $\rightarrow$ upperbound $\rightarrow$ xy $\rightarrow$ xby $\rightarrow$ sum $\rightarrow$ linear $\rightarrow$ ratio
  \item For linear constraints: position
  \item For constraints of the same type: x $\rightarrow$ y $\rightarrow$ z $\rightarrow$ w
  \item For constraints of the same type: $\leq$ $\rightarrow$ $\geq$
\end{enumerate}

\subsection{Data Augmentation}

\begin{figure}[ht!]
  \centering
  \includegraphics[width=1.0\linewidth]{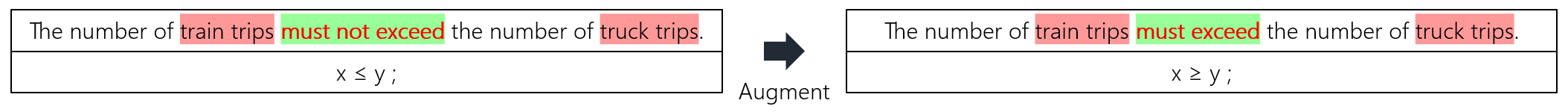}
  \caption{Data augmentation by reversing constraint direction.}
\end{figure}

I augment the data by reversing the direction of some constraints:

\begin{itemize}
  \item must not, can not, cannot $\rightarrow$ must
\end{itemize}

\section{Model}

I finetune $\textrm{BART}_\mathrm{large}$ \citep{BART}, a pretrained model for sequence-to-sequence tasks, for auto-formulation task. For a given input sequence $S = [w_1, w_2, ..., w_L]$, BART computes the token embeddings $E_{w_1}^{tok}, E_{w_2}^{tok}, ..., E_{w_L}^{tok} \in \mathbb{R}^d$ and position embeddings $E_1^{pos}, E_2^{pos}, ..., E_L^{pos} \in \mathbb{R}^d$, where $E^{tok}$ is the token embedding matrix, $E^{pos}$ is the position embedding matrix, and $d$ is the dimensionality of BART layers. Then, the sum of token embeddings and position embeddings (i.e., $E_{w_l}^{tok} + E_l^{pos}$) is forwarded to the BART encoder.

\begin{figure}[ht!]
  \centering
  \includegraphics[width=1.0\linewidth]{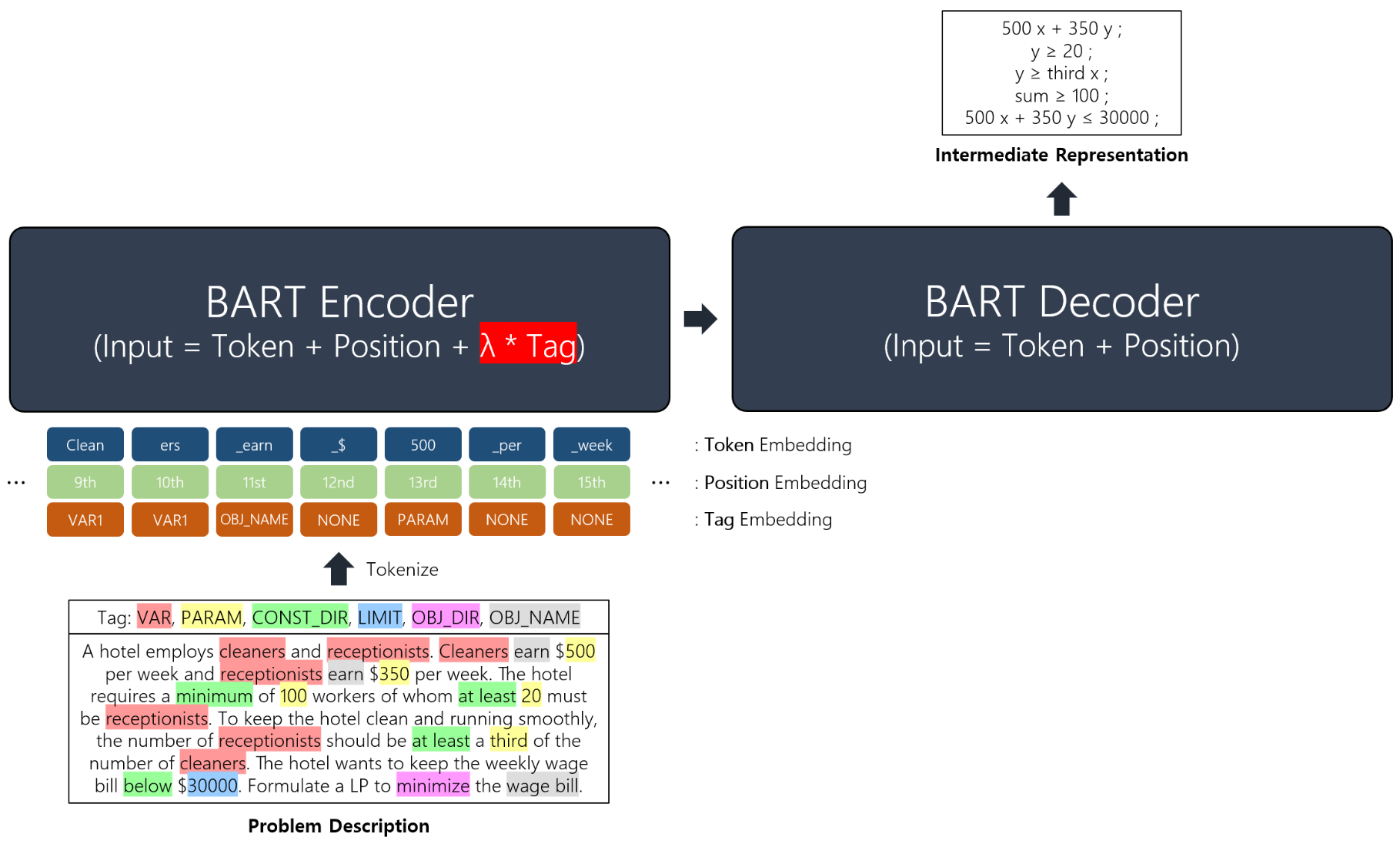}
  \caption{Method overview.}
\end{figure}

\subsection{Entity Tag Embedding}

I introduce entity tag embedding to utilize a given entity tag sequence. I compute the entity tag embeddings $E_{t_1}^{tag}, E_{t_2}^{tag}, ..., E_{t_L}^{tag}$ and add them to the input embeddings of the BART encoder (i.e., $E_{w_l}^{tok} + E_l^{pos} + E_{t_l}^{tag}$). To alleviate the destruction of pretrained knowledge, I initialize the tag embedding matrix $E^{tag}$ to 0s.

\subsection{Embedding Scaling}

When finetuning a pretrained model, the balance between the existing and newly introduced parameters has a significant impact on the accuracy of the model. To control this balance, we use an embedding scaling hyperparameter $\lambda$ for entity tag embedding. That is, the $l$-th input embedding of the BART encoder is $E_{w_l}^{tok} + E_l^{pos} + \lambda E_{t_l}^{tag}$.

\section{Experiments}

We use LPWP dataset \citep{LPWP}, a linear programming word problems dataset, to train and evaluate the model. To evaluate the accuracy of the model, we use declaration-level mapping accuracy defined as:

\begin{equation}
  \mathrm{Accuracy} = 1 - \frac{\sum_{i=1}^N \mathrm{FP}_i + \mathrm{FN}_i}{\sum_{i=1}^N \mathrm{D}_i},
\end{equation}

where N is the number of optimization problems, D is the number of ground truth declarations (i.e., objective and constraints), FP is the number of generated declarations that do not match the ground truth and FN is the number of ground truth declarations that the model failed to generate.

For model training, we use a batch size of 16, AdamW optimizer with a learning rate of 5e-5 and weight decay of 1e-5, and cosine annealing learning rate scheduler. We use gradient clipping with max norm of 1.0 and train the model for 100 epochs. For sequence generation, we use beam search with num beams of 4.

I released the code for the proposed method.\footnote{\url{https://github.com/jsh710101/nl4opt}}

\subsection{Ablation Study}

\begin{table}[ht!]
  \caption{The results of ablation study. $\lambda$ is the embedding scaling weight for entity tag and $p$ is the probability of reversing constraint direction for data augmentation.}
  \centering
  \begin{tabular}{ c c c c }
    \toprule
    \multicolumn{3}{c}{Hyperparameter} \\
    \cmidrule(r){1-3}
    BART Size & $\lambda$ & $p$ & Validation Accuracy \\
    \midrule
    Base  & 0 & 0   & 0.5513 \\
    Large & 0 & 0   & 0.7718 \\
    Large & 1 & 0   & 0.8000 \\
    Large & 5 & 0   & 0.8692 \\
    Large & 5 & 0.3 & 0.8846 \\
    \bottomrule
  \end{tabular}
\end{table}

The results of the ablation study demonstrate that 1) using a larger pretrained model, 2) adding entity tag embedding, 3) adjusting the weight of entity tag embedding, and 4) data augmentation improve the validation accuracy of the model. Surprisingly, just adjusting the scaling of entity tag embedding increased the validation accuracy by 8.65\%, which is a huge improvement considering the simplicity of this technique.

\subsection{Leaderboard}

\begin{table}[ht!]
  \caption{The leaderboard for NL4Opt competition subtask 2.}
  \centering
  \begin{tabular}{ c c }
    \toprule
    Team & Test Accuracy \\
    \midrule
    UIUC-NLP       & 0.899 \\
    Sjang          & 0.878 \\
    Long           & 0.867 \\
    PingAn-zhiniao & 0.866 \\
    Infrrd AI Lab  & 0.780 \\
    KKKKKi         & 0.634 \\
    \bottomrule
  \end{tabular}
\end{table}

The proposed method placed second in NL4Opt competition subtask 2.

\paragraph{Possible Improvement}

After analyzing the failure cases of the proposed method, I found that it often generate incorrect mathematical expressions for declarations described with previously unseen sentence structures. The accuracy of the model could be improved by using a larger pretrained model that has learned more sentence structures or by augmenting data considering diverse sentence structures.

\section{Conclusion}

I proposed a simple yet effective baseline for auto-formulation of problem description. Through the ablation study, I demonstrated the effectiveness of the proposed method, including entity tag embedding and well-defined intermediate representation, and it finally took second place in NeurIPS 2022 NL4Opt competition subtask 2. I believe that the proposed method has a lot of room for improvement, and I hope that researchers who study auto-formulation will find it useful.

\medskip

\bibliography{references}

\end{document}